\documentclass[]{article}
\newcommand{\ignore}[1]{}
\usepackage{amssymb}
\usepackage{amsmath}
\usepackage{epsfig}
\usepackage{colacl}

\newtheorem{theorem}{Theorem}
\newtheorem{definition}[theorem]{Definition}
\newtheorem{example}[theorem]{Example}
\newcommand{\I}{{\cal I}}

\newcommand{\sv}{{subj-verb}}

\title{\vspace*{-1.05in}
{\normalsize \em \hfill NAACL'00}\\
\vspace*{.55in}
A Classification Approach to Word Prediction\thanks{This 
research is supported by NSF grants IIS-9801638 and SBR-987345.} }
\author{Yair Even-Zohar  \hspace{0.8in}  Dan Roth \\ \\
  	 Department of Computer Science \\
         University of Illinois at Urbana-Champaign \\
         {\tt \{evenzoha,danr\}@uiuc.edu}}

\begin{document}

\maketitle

\begin{abstract}
The eventual goal of a language model is to accurately predict the
value of a missing word given its context. We present an approach to
word prediction that is based on learning a representation for each
word as a function of words and linguistics predicates in its
context. This approach raises a few new questions that we
address. 
First, in order to learn good word representations it is necessary to
use an expressive representation of the context. We present a way that
uses external knowledge to generate expressive context
representations, along with a learning method capable of handling the
large number of features generated this way that can, potentially,
contribute to each prediction.
Second, since the number of words ``competing'' for each prediction is
large, there is a need to ``focus the attention'' on a smaller subset
of these. We exhibit the contribution of a ``focus of attention''
mechanism to the performance of the word predictor. 
Finally, we describe a large scale experimental study in which the
approach presented is shown to yield significant improvements in word
prediction tasks.
\end{abstract}

\section{Introduction}


The task of predicting the most likely word based on properties of its
surrounding context is the archetypical prediction problem in natural
language processing (NLP). In many NLP tasks it is necessary
to determine the most likely word, part-of-speech (POS) tag or any
other token, given its history or context. Examples include part-of
speech tagging, word-sense disambiguation, speech recognition, accent
restoration, word choice selection in machine translation,
context-sensitive spelling correction and identifying discourse
markers.
Most approaches to these problems are based on \emph{n-gram}-like
modeling.  Namely, the learning methods make use of features which are
conjunctions of typically (up to) three consecutive words or POS tags in order
to derive the predictor.

In this paper we show that incorporating additional information
into the learning process is very beneficial. In particular, we
provide the learner with a rich set of features that combine the
information available in the local context along with shallow parsing
information. At the same time, we study a learning approach that is
specifically tailored for problems in which the potential number of
features is very large but only a fairly small number of them actually
participates in the decision.  Word prediction experiments that we
perform show significant improvements in error rate relative to the
use of the traditional, restricted, set of features.


\subsection*{Background}

The most influential problem in motivating statistical learning
application in NLP tasks is that of word selection in speech
recognition~\cite {Jelinek98}. There, word classifiers are derived
from a probabilistic language model which estimates the probability of
a sentence $s$ using Bayes rule as the product of conditional
probabilities,
\begin{eqnarray}
Pr(s) &\doteq &Pr(w_{1},w_{2},\ldots w_{n})=\notag \\ 
&\doteq &\Pi_{i=1}^{n}Pr(w_{i}|w_{1},\ldots w_{i-1}) \notag \label{eq:model} \\
&\doteq &\Pi _{i=1}^{n}Pr(w_{i}|h_{i})  \notag
\end{eqnarray}
where $h_{i}$ is the relevant \emph{history} when predicting $w_{i}$. 
Thus, in order to predict the most likely word in a given context, a
global estimation of the sentence probability is derived which, in
turn, is computed by estimating the probability of each word given its
local context or history. Estimating terms of the form $Pr(w|h)$ is
done by assuming some generative probabilistic model, typically using
Markov or other independence assumptions, which gives rise to
estimating conditional probabilities of n-grams type features (in the
word or POS space). Machine learning based classifiers and maximum
entropy models which, in principle, are not restricted to features of
these forms have used them nevertheless, perhaps under the influence
of probabilistic methods~\cite{Brill95,Yarowsky94,RRR94}.

It has been argued that the information available in the local context
of each word should be augmented by global sentence information and
even information external to the sentence in order to learn better
classifiers and language models.
Efforts in this directions consists of (1) directly adding syntactic
information, as in~\cite{ChelbaJe98,Rosenfeld96}, and (2) indirectly
adding syntactic and semantic information, via similarity models; in this case n-gram type features
are used whenever possible, and when they cannot be used (due to data
sparsity), additional information compiled into a similarity
measure is used~\cite {DaganLePe99}. Nevertheless, the efforts in this
direction so far have shown very insignificant improvements, if
any~\cite{ChelbaJe98,Rosenfeld96}. We believe that the main reason for
that is that incorporating information sources in NLP needs to be
coupled with a learning approach that is suitable for it.

Studies have shown that both machine learning and probabilistic
learning methods used in NLP make decisions using a linear decision
surface over the feature space \cite{Roth98,Roth99p}.
In this view, the feature space consists of simple functions (e.g.,
n-grams) over the the original data so as to allow for expressive
enough representations using a simple functional form (e.g., a linear
function).
This implies that the number of potential features that the learning
stage needs to consider may be very large, and may grow rapidly when
increasing the expressivity of the features.
Therefore a feasible computational approach needs to be
feature-efficient. It needs to tolerate a large number of potential
features in the sense that the number of examples required for it to
converge should depend mostly on the number features relevant to the
decision, rather than on the number of potential features.

This paper addresses the two issues mentioned above. It presents a
rich set of features that is constructed using information readily
available in the sentence along with shallow parsing and dependency
information. It then presents a learning approach that can use this
expressive (and potentially large) intermediate representation and
shows that it yields a significant improvement in word error rate for
the task of word prediction.


The rest of the paper is organized as follows. In
section~\ref{sec:features} we formalize the problem, discuss the
information sources available to the learning system and how we use
those to construct features. In section~\ref{sec:learning} we present
the learning approach, based on the SNoW learning architecture.
Section~\ref{sec:experiments} presents our experimental study
and results. In
section~\ref{sec:focus} we discuss the issue of deciding on a set of
candidate words for each decision.
Section~\ref{sec:conc} concludes and discusses future work.


\section{Information Sources and Features}
\label{sec:features}\label{features}

Our goal is to learn a representation for each word in terms of
features which characterize the syntactic and semantic context in
which the word tends to appear.
Our features are defined as simple relations over a collection of
predicates that capture (some of) the information available in a
sentence.
\subsection{Information Sources}
\begin{definition}
Let $s=<w_{1},w_{2},...,w_{n}>$ be a sentence in which $w_{i}$ is 
the $i$-th word. 
Let $\I$ be a collection of predicates over a sentence $s$.
IS(s))\footnote{We denote IS(s) as IS wherever it is obvious what
the referred sentence we is, or whenever we want to indicate
Information Source in general.}, 
the \textbf{Information source(s)} available for the sentence $s$ is a
representation of $s$ as a list of predicates $I \in \I$,
$$IS(s)=\{I_{1}(w_{1_1},...w_{1_i}),...,I_{k}(w_{k_1},...w_{k_i})\}.$$
$j_i$ is the arity of the predicate $I_j$.
\end{definition}

\begin{example}\label{ex:IS}
Let $s$ be the sentence \\
$<{\tt John, X, at, the, clock, to, see, what, time, it, is}>$
Let $\I$=\{word, pos, \sv\}, with the interpretation that
{\tt word} is a unary predicate that returns the value of the word
in its domain;
{\tt pos} is a unary predicate that returns the value of the pos of
the word in its domain, in the context of the sentence;
{\tt $\sv$} is a binary predicate that returns the value of the 
two words in its domain if the second is a verb in the sentence and the first
is its subject; it returns $\phi$ otherwise. 
Then,
\begin{eqnarray*}
IS(s) & = & \{ word(w_{1}) = {\tt John},...,
               word(w_{3}) = {\tt at},..., \\
       &   & word(w_{11}) = {\tt is}, 
               pos(w_{4}) = DET,..., \\
       &   & \sv(w_1,w_2) = \{{\tt John,X}\}...\}.
\end{eqnarray*}

\end{example}
The IS representation of $s$ consists only of the predicates with
non-empty values. E.g., {\tt $pos(w_6)=modal$} is not part of the IS for
the sentence above. $\sv$ might not exist at all in the IS even if the
predicate is available, e.g., in {\tt The ball was given to Mary}.

Clearly the $IS$ representation of $s$ does not contain all the
information available to a human reading $s$; it captures, however, 
all the input that is available to the computational process discussed
in the rest of this paper.  The predicates could be generated by any
external mechanism, even a learned one. This issue is orthogonal
to the current discussion.

\subsection{Generating Features}

Our goal is to learn a representation for each word of interest. Most
efficient learning methods known today and, in particular, those used
in NLP, make use of a linear decision surface over their feature
space~\cite{Roth98,Roth99p}. Therefore, in order to learn expressive
representations one needs to compose complex features as a function of
the information sources available. A linear function expressed directly in terms
of those will not be expressive enough.
We now define a language that allows one to define ``types'' of
features\footnote{We note that we do not define the features will be used 
in the learning process. These are going to be
defined in a data driven way given the definitions discussed here and
the input ISs.  The importance of formally defining the ``types'' is
due to the fact that some of these are quantified. Evaluating them on
a given sentence might be computationally intractable and a formal
definition would help to flesh out the difficulties and aid in designing
the language~\cite{CumbyRo00}.} in terms of the information sources
available to it.
%
%

\begin{definition}[Basic Features] 
Let $I \in \I$ be a $k$-ary predicate with range $R$.
Denote $w^k = (w_{j_{1}},\ldots,w_{j_{k}})$.
We define two basic binary relations as follows. For $\alpha \in R$ we define:
\begin{equation}
f(I(w^k),\alpha )=\left\{ 
\begin{array}{l}
1\text{ iff }I(w^k)=\alpha\\ 
0\text{ otherwise}
\end{array}
\right. 
\end{equation}
An existential version of the relation is defined by:
\begin{equation}
f(I(w^k), x)=\left\{ 
\begin{array}{l}
1\text{ iff } \exists \alpha \in R \, s.t \, I(w^k)=\alpha\\ 
0\text{ otherwise}
\end{array}
\right. 
\end{equation}
\end{definition}
Features, which are defined as binary relations, can be composed to
yield more complex relations in terms of the original predicates
available in IS.
\begin{definition} [Composing features]
Let $f_1,f_2$ be feature definitions. Then $f_{and}(f_1,f_2)$
$f_{or}(f_1,f_2)$ $f_{not}(f_1)$ are defined and given the usual semantic:

\begin{eqnarray}
f_{and}(f_{1},f_{2}) &=&\left\{ 
\begin{array}{l}
1\text{ if }f_{1}=f_{2}=1 \\ 
0\text{ otherwise}
\end{array}
\right. \notag  \\
f_{or}(f_{1},f_{2}) &=&\left\{ 
\begin{array}{l}
1\text{ if }f_{1}=1\text{ or }f_{2}=1 \\ 
0\text{ otherwise}
\end{array}
\right.  \notag  \\
f_{not}(f_{1}) &=&\left\{ 
\begin{array}{l}
1\text{ if }f_{1}=0 \\ 
0\text{ otherwise}
\end{array}
\right.  \notag  
\end{eqnarray}
\end{definition}
In order to learn with features generated using these definitions as
input, it is important that features generated when applying the
definitions on different ISs are given the same identification.  In
this presentation we assume that the composition operator
along with the appropriate IS element (e.g., Ex.~\ref{ex:IS},
Ex.~\ref{ex:linear}) are written explicitly as the identification of
the features.  Some of the subtleties in defining the output
representation are addressed in~\cite{CumbyRo00}.

\subsection{Structured Features}
So far we have presented features as relations over $IS(s)$ and
allowed for Boolean composition operators. In most cases more
information than just a list of active predicates is available.  We
abstract this using the notion of a {\em structural information
source $(SIS(s))$} defined below.  This allows richer
class of feature types to be defined. 

\subsection{Structured Instances}
\begin{definition} [Structural Information Source]
Let $s=<w_{1},w_{2},...,w_{n}>$.
SIS(s)), the {\bf Structural Information source(s)} available for the
sentence $s$, is a tuple $(s,E_1,\ldots, E_k)$ of directed acyclic
graphs with $s$ as the set of vertices and $E_i$'s, a set of edges in $s$.
\end{definition}

\begin{example}[Linear Structure]
The simplest SIS is the one corresponding to the linear structure of
the sentence.  That is, $SIS(s) = (s,E)$ where $(w_i,w_j) \in E$ iff
the word $w_i$ occurs immediately before $w_j$ in the sentence
(Figure~\ref{fig_dependency_example} bottom left part).
\end{example}
In a linear structure $(s=<w_{1},w_{2},...,w_{n}>,E)$, where $E
= \{(w_i,w_{i+1});i=1,\ldots n-1\}$, we define the chain
$$ C(w_j,[l,r]) = \{w_{j-l},\ldots,w_j,\ldots w_{j+r}\}
\cap s.$$

We can now define a new set of features that makes use of the
structural information.  Structural features are defined using the
SIS. When defining a feature, the naming of nodes in $s$ is done
relative to a distinguished node, denoted $w_p$, which we call the
{\em focus word} of the feature. 
Regardless of the arity of the
features we sometimes denote the feature $f$ defined with respect to
$w_p$ as $f(w_p)$.
\begin{definition}[Proximity]\label{def:prox}
Let $SIS(s)=(s,E)$ be the linear structure and let $I \in \I$ be a
$k$-ary predicate with range $R$.  Let $w_p$ be a focus word and 
$C=C(w_p,[l,r])$ the chain around it.  Then, the proximity
features for $I$ with respect to the chain $C$ are defined as:
\begin{equation}
f_{C}(I(w),\alpha )=\left\{ 
\begin{array}{l}
1\text{ if }I(w)=\alpha, \alpha \in R, w \in C \\
0\text{ otherwise}
\end{array}
\right. 
\end{equation}
\end{definition}

The second type of feature composition defined using the
structure is a collocation operator.
\begin{definition}[Collocation] \label{def:colloc}
Let $f_1,\ldots f_k$ be feature definitions. $colloc_{C}(f_1,f_2,\ldots
f_k)$ is a restricted conjunctive operator that is evaluated on a
chain $C$ of length $k$ in a graph. 
Specifically, let $C = \{w_{j_1},w_{j_2},\ldots,w_{j_k}\}$
be a chain of length $k$ in $SIS(s)$. Then, the collocation 
feature for $f_1,\ldots f_k$ with respect to the chain $C$ is defined as

\begin{equation}
colloc_{C}(f_1,\ldots,f_k)=\left\{ 
\begin{array}{l}
1\text{ if } \forall i=1,\ldots k, f_i(w_{j_i}) =1 \\
0\text{ otherwise}
\end{array}
\right. 
\end{equation}
\end{definition}

The following example defines features that are used in the experiments 
described in Sec.~\ref{sec:experiments}.
\begin{example} \label{ex:linear}
Let $s$ be the sentence in Example \ref{ex:IS}.  We define some of the features
with respect to the linear structure of the sentence.
The word $X$ is used as the focus word and a chain $[-10,10]$ is
defined with respect to it. The proximity features are defined with
respect to the predicate {\tt word}. We get, for example: ${\tt
f_{C}(word)= John; f_{C}(word)= at; f_{C}(word)= clock.}$

Collocation features are defined with respect to a chain $[-2,2]$
centered at the focus word $X$.  They are defined with respect to two
basic features $f_1,f_2$ each of which can be either $f(word,\alpha)$
or $f(pos,\alpha)$.  The resulting features include, for example:
\begin{eqnarray*}
     colloc_{C}(word,word) & = & \{John-X\}; \\
     colloc_{C}(word,word) & = & \{X-at\}; \\
     colloc_{C}(word,pos) & = & \{at-DET\}.
\end{eqnarray*}
\end{example}

\subsection{Non-Linear Structure}

\begin{figure*}
\hspace {-0.1in}
\epsfig{file=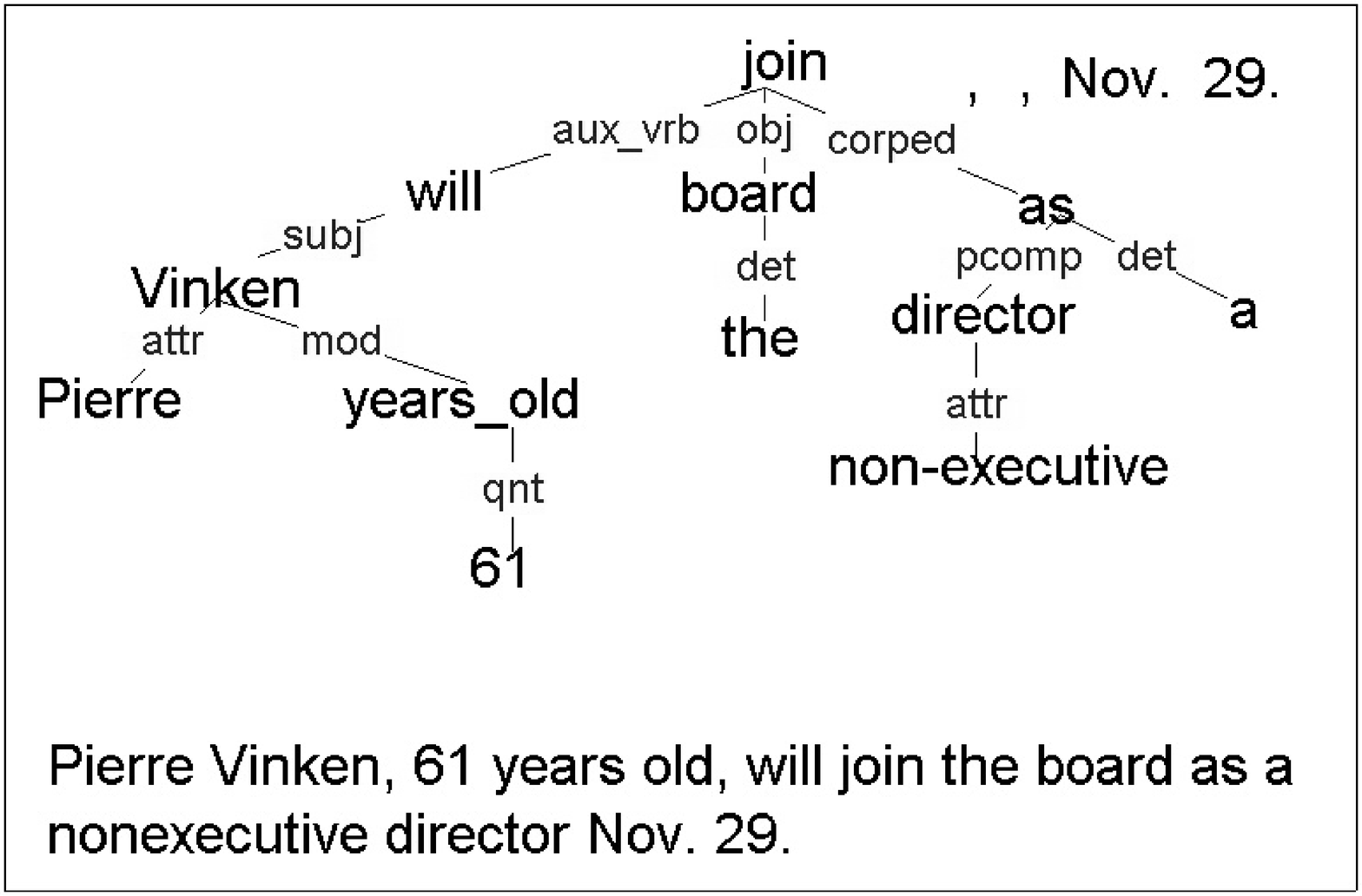,height=2.0695in,width=3.4481in}
\hspace {-0.1in}
\epsfig{file=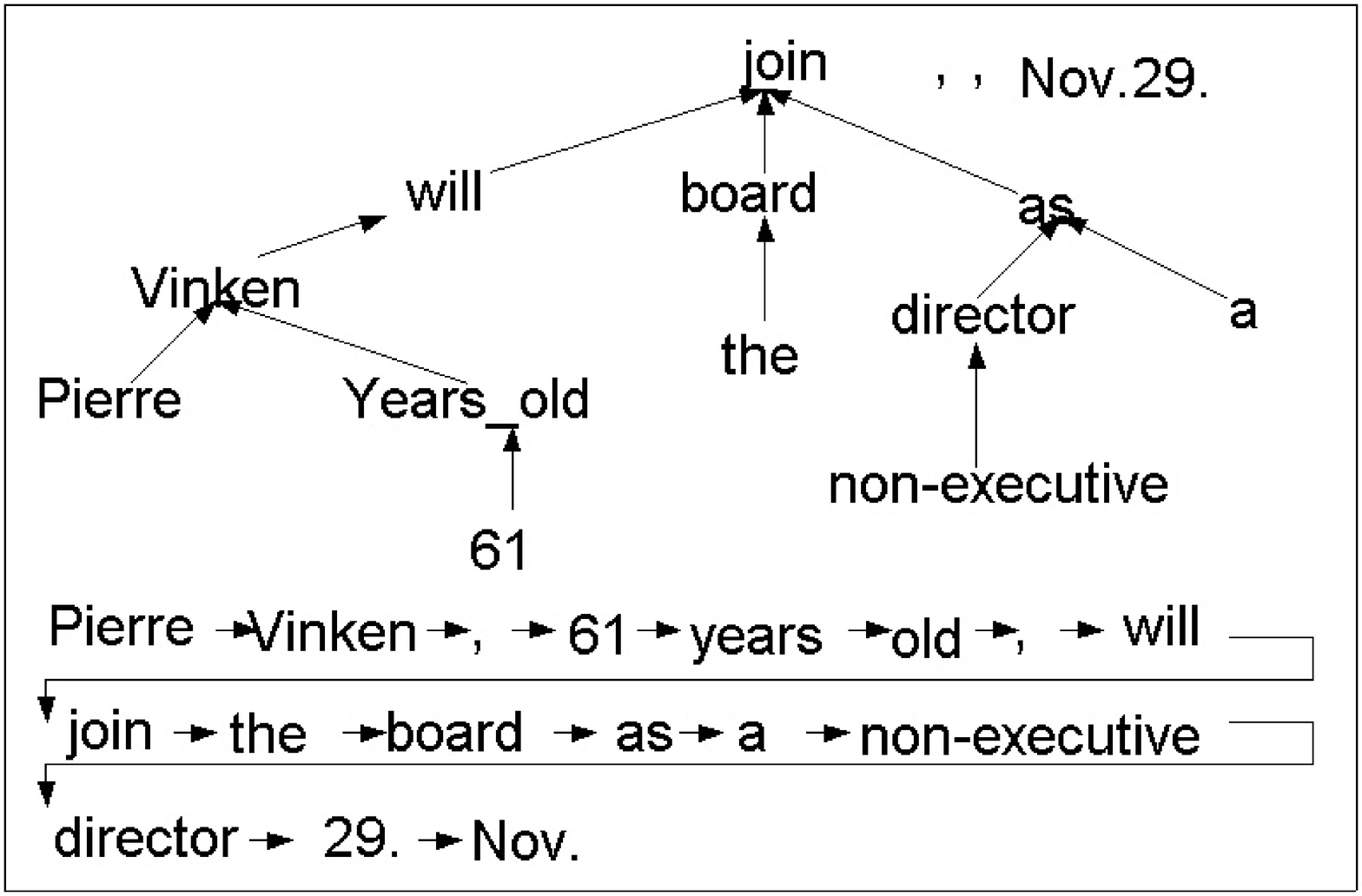,height=2.0695in,width=3.4481in}
\caption{\textbf{A sentence with a linear and a dependency grammar structure}
\label{fig_dependency_example}}
\end{figure*}
So far we have described feature definitions which make use of the
linear structure of the sentence and yield features which are not too
different from standard features used in the literature e.g., $n$-grams with
respect to $pos$ or $word$ can be defined as $colloc$ for the appropriate
chain. Consider now
that we are given a general directed acyclic graph $G=(s,E)$ on the the
sentence $s$ as its nodes. Given a distinguished {\em focus word} $w_p
\in s$ we can define a chain in the graph as we did above for the
linear structure of the sentence.
Since the definitions given above, Def.~\ref{def:prox} and
Def.~\ref{def:colloc}, were given for chains they would apply for any
chain in any graph. This generalization becomes interesting if we are
given a graph that represents a more involved structure of the sentence.

Consider, for example the graph $DG(s)$ in
Figure~\ref{fig_dependency_example}.  
$DG(s)$ described the dependency graph of the sentence $s$.  An edge
$(w_{i},w_{j})$ in $DG(s)$ represent a dependency between the two
words. In our feature generation language we separate the information
provided by the dependency grammar\footnote{This information can be
produced by a functional dependency grammar (FDG), which assigns each
word a specific function, and then structures the sentence
hierarchically based on it, as we do here~\cite{TapanainenJa97}, but
can also be generated by an external rule-based parser or a learned
one.}
to two parts. The structural information, provided in the left side of
Figure~\ref{fig_dependency_example}, is used to generate $SIS(s)$. The
labels on the edges are used as predicates and are part of $IS(s).$
Notice that some authors~\cite{Yuret:PhD98,BergerPr98} have used the
structural information, but have not used the information given by the
labels on the edges as we do.

The following example defines features that are used in the experiments 
described in Sec.~\ref{sec:experiments}.
\begin{example} \label{ex:nonlinear}
Let $s$ be the sentence in Figure~\ref{fig_dependency_example} along
with its IS that is defined using the predicates {\tt word, pos, subj,
obj, aux\_vrb}.
A {\tt subj-verb} feature, $f_{subj-verb}$, can be defined as a
collocation over chains constructed with respect to the focus word
{\tt join}. Moreover, we can define $f_{subj-verb}$ to be active also
when there is an {\tt aux\_vrb} between the {\tt subj} and {\tt verb},
by defining it as a disjunction of two collocation features,
the {\tt subj-verb} and the {\tt subj-aux\_vrb-verb}.
Other features that we use are conjunctions of words that occur before
the focus verb (here: {\tt join}) along all the chains it occurs in
(here: {\tt will, board, as}) and collocations of {\tt obj} and {\tt
verb}.
\end{example}

As a final comment on feature generation, we note that the language
presented is used to define ``types'' of features. These are
instantiated in a data driven way given input sentences. A large
number of features is created in this way, most of which might
not be relevant to the decision at hand; thus, this process needs to be
followed by a learning process that can learn in the presence of these
many features.

\section{The Learning Approach\label{Learning_approach}\label{sec:learning}}

Our experimental investigation is done using the \emph{SNoW} learning
system \cite{Roth98}. Earlier versions of \emph{SNoW} \cite
{Roth98,GoldingRo99,RothZe98,MPRZ99} have been applied successfully to
several natural language related tasks. Here we use \emph{SNoW} for
the task of word prediction; a representation is learned for each word
of interest, and these compete at evaluation time to determine the
prediction.

\subsection{The \emph{SNOW}\ Architecture}
The \emph{SNoW} architecture is a sparse network of linear units over
a common pre-defined or incrementally learned feature space. It is
specifically tailored for learning in domains in which the potential
number of features might be very large but only a small subset of them
is actually relevant to the decision made.

Nodes in the input layer of the network represent simple relations on
the input sentence and are being used as the input features. Target
nodes represent words that are of interest; in the case studied here,
each of the word candidates for prediction is represented as a target
node.
An input sentence, along with a designated word of interest in it, is
mapped into a set of features which are active in it; this
representation is presented to the input layer of SNoW and propagates
to the target nodes.
Target nodes are linked via weighted edges to (some of) the input features.
Let $\mathcal{A}_{t} = \{i_1, \ldots, i_m \}$ be the set of features that are
active in an example and are linked to the target node $t$. Then the linear
unit corresponding to $t$ 
is \emph{active} iff 
$$\sum_{i \in \mathcal{A}_{t}} w^t_i > \theta_t,$$
where $w^t_i$ is the weight on the edge connecting the $i$th feature to the
target node $t$, and $\theta_t$ is the threshold for the target node $t$.
In this way, \emph{SNoW} provides a collection of \emph{word
representations} rather than just discriminators.

A given example is treated autonomously by each target subnetwork; an
example labeled $t$ may be treated as a positive example by the subnetwork
for $t$ and as a negative example by the rest of the target nodes.
The learning policy is on-line and mistake-driven; several update
rules can be used within SNoW. The most successful update rule 
is a variant of Littlestone's Winnow
update rule~\cite{Littlestone88}, a multiplicative update rule that is
tailored to the situation in which the set of input features is not
known a priori, as in the infinite attribute model~\cite{Blum92}.
This mechanism is implemented via the sparse architecture of SNoW. That is,
(1) input features are allocated in a data driven way -- an input node for the
feature $i$ is allocated only if the feature $i$ was active in any input
sentence and (2) a link (i.e., a non-zero weight) exists between a target node
$t$ and a feature $i$ if and only if $i$ was active in an example labeled $t$.

One of the important properties of the sparse architecture is that the
complexity of processing an example depends only on the number of
features active in it, $n_a$, and is independent of the total number of
features, $n_t$, observed over the life time of the system. This is
important in domains in which the total number of features is very
large, but only a small number of them is active in each example.

Once target subnetworks have been learned and the network is being
evaluated, a decision support mechanism is employed, which selects the
dominant active target node in the SNoW unit via a winner-take-all
mechanism to produce a final prediction. 
SNoW is available publicly at {\tt
http://L2R.cs.uiuc.edu/\~{}cogcomp.html}.

\section{Experimental Study\label{experiment}\label{sec:experiments}}

\subsection{Task definition$\label{Task Definition}$}

The experiments were conducted with four goals in mind:

\begin{enumerate}
\item To compare mistake driven algorithms with naive Bayes, trigram
with backoff and a simple maximum likelihood estimation (MLE) baseline.

\item  To create a set of experiments which is comparable with similar
experiments that were previously conducted by other researchers.

\item To build a baseline for two types of extensions of the simple
use of linear features: (i) Non-Linear features (ii) Automatic focus
of attention.

\item To evaluate word prediction as a simple language model.
\end{enumerate}

We chose the verb prediction task which is similar to other word
prediction tasks (e.g.,\cite{GoldingRo99}) and, in particular, follows
the paradigm in~\cite{LeePe99b,DaganLePe99,Lee99}.
There, a list of the confusion sets is constructed first, each consists
of two different verbs. The verb $v_{1\text{ }}$ is coupled
with $v_{2}$ provided that they occur equally likely in the corpus.
In the test set, every occurrence of $v_{1}$ or $v_{2}$ was replaced by
a set $\{v_{1},v_{2}\}$ and the classification task was to predict the
correct verb. For example, if a confusion set is created for the verbs
''make'' and ''sell'', then the data is altered as follows:
\begin{eqnarray*}
\text{make the paper} &\rightarrow &\text{\{make,sell\} \, the paper
} \\
\text{sell sensitive data} &\rightarrow &\text{\{make,sell\} \, sensitive
data
}
\end{eqnarray*}
The  evaluated predictor chooses which of the two verbs is more
likely to occur in the current sentence.

In choosing the prediction task in this way, we make sure the task in
difficult by choosing between competing words that have the same prior
probabilities and have the same part of speech.
A further advantage of this paradigm is that in future experiments we
may choose the candidate verbs so that they have the same
sub-categorization, phonetic transcription, etc. in order to imitate
the first phase of language modeling used in creating candidates for
the prediction task.
Moreover, the pre-transformed data provides the correct answer so
that (i) it is easy to generate training data; no supervision is
required, and (ii) it is easy to evaluate the results assuming that
the most appropriate word is provided in the original text.

Results are evaluated using word-error rate
(WER). Namely, every time we predict the wrong word it is counted as a
mistake.

\subsection{Data}

We used the Wall Street Journal (WSJ) of the years 88-89. The size of
our corpus is about 1,000,000 words. The corpus was divided into 80\%
training and 20\% test. The training and the test data were processed
by the FDG parser \cite{TapanainenJa97}. 
Only verbs that occur at least 50 times in the corpus
were chosen. This resulted in $278$ verbs that we split into $139$
confusion sets as above. 
After filtering the examples of verbs which were not in any of
the sets we use $73,184$ training examples and $19,852$ test examples.

\subsection{Results}

\subsubsection{Features\label{Local Features}}

In order to test the advantages of different feature sets we conducted
experiments using the following features sets:

\begin{enumerate}
\item  Linear features: proximity of window size $\pm 10$ words, conjunction
of size 2 using window size $\pm 2$. The conjunction combines words and
parts of speech.

\item Linear + Non linear features: using the linear features defined
in (1) along with non linear features that use the predicates {\tt
subj, obj, word, pos}, the collocations {\tt subj-verb}, {\tt verb-obj}
linked to the focus verb via the graph structure and conjunction of
$2$ linked words.
\end{enumerate}
The over all number of features we have generated for all $278$ target
verbs was around $400,000$. In all tables below the NB columns
represent results of the naive Bayes algorithm as implemented
within SNoW and the SNoW column represents the results of the sparse
Winnow algorithm within SNoW.

\begin{table}[h] \centering
\begin{tabular}{|l|l|l|l|}
\hline
& \textbf{Bline} & \textbf{NB} & \textbf{SNoW} \\ \hline
\textbf{Linear} & $49.6$ & $13.54$ & $11.56$ \\ \hline
\textbf{Non Linear} & $49.6$ & $12.25$ & $9.84$ \\ \hline
\end{tabular}
\caption{\bf Word Error Rate results for linear and non-linear features}
\label{table_local_WER}
\end{table}

Table \ref{table_local_WER} summarizes the results of the experiments
with the features sets (1), (2) above.  The baseline experiment uses
MLE, the majority predictor.  In addition, we conducted the same
experiment using trigram with backoff and the WER is $29.3\%$. From
these results we conclude that using more expressive features helps
significantly in reducing the WER.  However, one can use those types of
features only if the learning method handles large number of possible
features. This emphasizes the importance of the new learning method.

\begin{table}[h] \centering
\begin{tabular}{|l|l|l|l|}
\hline
& \textbf{Similarity} & \textbf{NB} & \textbf{SNoW} \\ \hline
\textbf{WSJ data} &  & $54.6\%$ & $59.1\%$ \\ \hline
\textbf{AP news} & $47.6\%$ &  &  \\ \hline
\end{tabular}
\caption{\bf Comparison of the improvement achieved using similarity
methods~\cite{DaganLePe99} and using the methods presented in this
paper. Results are shown in percentage of improvement in accuracy over the
baseline.}
\label{table_similarity_WER}
\end{table}

Table \ref{table_similarity_WER} compares our method to methods that
use similarity measures \cite{DaganLePe99,Lee99}. Since we could not
use the same corpus as in those experiments, we compare the ratio of
improvement and not the WER.  The baseline in this studies is
different, but other than that the experiments are identical. We show
an improvement over the best similarity method. Furthermore, we train
using only $73,184$ examples while \cite {DaganLePe99} train using
$587,833$ examples.  Given our experience with our approach on other
data sets we conjecture that we could have improved the results
further had we used that many training examples.
%

\subsection{Focus of attention}\label{sec:focus}

SNoW is used in our experiments as a multi-class predictor - a
representation is learned for each word in a given set and, at
evaluation time, one of these is selected as the prediction. The set
of candidate words is called the {\em confusion
set}~\cite{GoldingRo99}.
Let $C$ be the set of all target words.  In previous experiments we
generated artificially subsets of size $2$ of $C$ in order to evaluate
the performance of our methods. In general, however, the question of
determining a good set of candidates is interesting in it own right. 
In the absence of a good method, one might end up choosing a verb from among a larger
set of candidates. We would like to study the effects this issue has
on the performance of our method.

In principle, instead of working with a single large confusion set
$C$, it might be possible to split $C$ into subsets of smaller
size. This process, which we call the {\em focus of attention (FOA)}
would be beneficial only if we can guarantee that, with high
probability, given a prediction task, we know which confusion set to
use, so that the true target belongs to it.
In fact, the FOA problem can be discussed separately for the training
and test stages.

\begin{enumerate}
\item  Training: 
Given our training policy (Sec.~\ref{sec:learning}) every positive
example serves as a negative example to all other targets in its confusion
set.  For a large set $C$ training might become computationally
infeasible.

\item Testing: considering only a small set of words as candidates at
evaluation time increases the baseline and might be significant from
the point of view of accuracy and efficiency.
\end{enumerate}
To evaluate the advantage of reducing the size of the confusion set in
the training and test phases, we conducted the following
experiments using the same features set (linear features as in Table
\ref{table_local_WER}).

\begin{table}[h] \centering
\begin{tabular}{|l|l|l|l|}
\hline
& \textbf{Bline} & \textbf{NB} & \textbf{SNoW} \\ \hline
\textbf{Train All Test All} & $87.44$ & $65.22$ & $65.05$ \\ \hline
\textbf{Train All Test 2} & $49.6$ & $13.54$ & $13.15$ \\ \hline
\textbf{Train 2 Test 2} & $49.6$ & $13.54$ & $11.55$ \\ \hline
\end{tabular}
\caption{\bf Evaluating Focus of Attention: Word Error Rate for
Training and testing using all the words together against using pairs
of words.}
\label{table_FOA}
\end{table}
``Train All'' means training on all 278 targets together. ``Test all''
means that the confusion set is of size 278 and includes all the
targets.  The results shown in Table \ref{table_FOA} suggest that, in
terms of accuracy, the significant factor is the confusion set size
in the test stage. 
The effect of the confusion set size on training is minimal (although
it does affect training time).  We note that for the naive Bayes
algorithm the notion of negative examples does not exist, and
therefore regardless of the size of confusion set in training, it
learns exactly the same representations. Thus, in the
NB column, the confusion set size in training makes no
difference.

The application in which a word predictor is used might give a partial
solution to the FOA problem.  For example, given a prediction task in
the context of speech recognition the phonemes that constitute the word
might be known and thus suggest a way to generate a small confusion
set to be used when evaluating the predictors.

Tables \ref{table_speech1},\ref{table_speech2} present the results of
using artificially simulated speech recognizer using a method of
general phonetic classes.  That is, instead of transcribing a word by
the phoneme, the word is transcribed by the phoneme {\em
classes}\cite{JurafskyMa00}.
Specifically, these experiments deviate from the task definition given
above. The confusion sets used are of different sizes and they consist
of verbs with different prior probabilities in the corpus.
Two sets of experiments were conducted that use the phonetic
transcription of the words to generate confusion sets.

\begin{table}[h] \centering

\begin{tabular}
[c]{|l|l|l|l|}\hline
& \textbf{Bline} & \textbf{NB} & \textbf{SNoW}\\\hline
\textbf{Train All Test PC} & 19.84 & 11.6 & 12.3\\\hline
\textbf{Train PC Test PC} & 19.84 & 11.6 & 11.3\\\hline
\end{tabular}
\caption{\bf Simulating Speech Recognizer: Word Error Rate for
Training and testing with confusion sets determined based on phonetic
classes (PC) from a simulated speech recognizer.}
\label{table_speech1}
\end{table}

In the first experiment (Table \ref{table_speech1}), the transcription
of each word is given by the broad phonetic groups to which the phonemes
belong i.e., nasals, fricative, etc.\footnote{In this experiment, the
vowels phonemes were divided into two different groups to account for
different sounds.}. For example, the word ''b\_u\_y'' is transcribed
using phonemes as ''b\_Y'' and here we transcribe it as ''P\_V1''
which stands for ''Plosive\_Vowel1''. This partition results in a
partition of the set of verbs into several confusions sets. A few of
these confusion sets consist of a single word and therefore have 100\%
baseline, which explains the high baseline.

\begin{table}[here] \centering
\begin{tabular}
[c]{|l|l|l|l|}\hline
& \textbf{Bline} & \textbf{NB} & \textbf{SNoW}\\\hline
\textbf{Train All Test PC} & 45.63 & 26.36 & 27.54\\\hline
\textbf{Train PC Test PC} & 45.63 & 26.36 & 25.55\\\hline
\end{tabular}
\caption{\bf Simulating Speech Recognizer: Word Error Rate for 
Training and testing with confusion sets determined based on phonetic
classes (PC) from a simulated speech recognizer. In this case
only confusion sets that have less than 98\% baseline are used, which explains
the overall lower baseline.}
\label{table_speech2}
\end{table}

Table \ref{table_speech2} presents the results of a similar experiment
in which only confusion sets with multiple words were used, resulting
in a lower baseline.

As before, Train All means that training is done with all 278 targets 
together while Train PC 
means that the PC confusion sets were used also in training.
We note that for the case of SNoW, used here with the sparse Winnow
algorithm, that size of the confusion set in training has some,
although small, effect.  The reason is that when the training is done
with all the target words, each target word representation 
with all the examples in which it does not occur are used as negative
examples. When a smaller confusion set is used the negative examples
are more likely to be ``true'' negative.

\section{Conclusion\label{sec:conc}}

This paper presents a new approach to word prediction tasks. For each
word of interest, a word representation is learned as a function of a
common, but potentially very large set of expressive (relational)
features. Given a prediction task (a sentence with a missing word)
the word representations are evaluated on it and compete for the most
likely word to complete the sentence.

We have described a language that allows one to define expressive
feature types and have exhibited experimentally the advantage of
using those on word prediction task. We have argued that the success
of this approach hinges on the combination of using a large set of
expressive features along with a learning approach that can tolerate
it and converges quickly despite the large dimensionality of the data.
We believe that this approach would be useful for other disambiguation 
tasks in NLP. 

We have also presented a preliminary study of a reduction in the
confusion set size and its effects on the prediction performance. In
future work we intend to study ways that determine the appropriate
confusion set in a way to makes use of the current task properties.

\subsubsection*{Acknowledgments}
We gratefully acknowledge helpful comments and programming help
from Chad Cumby. 
\bibliographystyle{acl00}
\bibliography{learn,nlp}

\end{document}